\documentclass[a4paper,english,conference]{IEEEtran}
\usepackage[T1]{fontenc}
\usepackage[latin9]{inputenc}
\usepackage{verbatim}
\usepackage{amsmath}
\usepackage{amssymb}
\usepackage{graphicx}

\makeatletter

\providecommand{\tabularnewline}{\\}

\makeatother

\usepackage{babel}
\begin{document}

\title{Enhanced Robot Audition Based on Microphone Array Source Separation
with Post-Filter}

\author{\authorblockN{Jean-Marc Valin, Jean Rouat, Fran\c{c}ois Michaud}\authorblockA{LABORIUS, Department of Electrical Engineering and Computer Engineering\\
Universit\'e de Sherbrooke, Sherbrooke (Quebec) CANADA, J1K 2R1\\
\{Jean-Marc.Valin, Jean.Rouat, Francois.Michaud\}@USherbrooke.ca}}
\maketitle
\begin{abstract}
\footnotetext{\copyright 2004 IEEE.  Personal use of this material is permitted. Permission from IEEE must be obtained for all other uses, in any current or future media, including reprinting/republishing this material for advertising or promotional purposes, creating new collective works, for resale or redistribution to servers or lists, or reuse of any copyrighted component of this work in other works.}We
propose a system that gives a mobile robot the ability to separate
simultaneous sound sources. A microphone array is used along with
a real-time dedicated implementation of Geometric Source Separation
and a post-filter that gives us a further reduction of interferences
from other sources. We present results and comparisons for separation
of multiple non-stationary speech sources combined with noise sources.
The main advantage of our approach for mobile robots resides in the
fact that both the frequency-domain Geometric Source Separation algorithm
and the post-filter are able to adapt rapidly to new sources and non-stationarity.
Separation results are presented for three simultaneous interfering
speakers in the presence of noise. A reduction of log spectral distortion
(LSD) and increase of signal-to-noise ratio (SNR) of approximately
10~dB and 14~dB are observed.
\end{abstract}

\section{Introduction}

Our hearing sense allows us to perceive all kinds of sounds (speech,
music, phone ring, closing a door, etc.) in our world, whether we
are moving or not. To operate in human and natural settings, autonomous
mobile robots should be able to do the same. This requires the robots
not just to detect sounds, but also to localise their origin, separate
the different sound sources (since sounds may occur simultaneously),
and process all of this data to extract useful information about the
world.

Even though artificial hearing would be an important sensing capability
for autonomous systems, the research topic is still in its infancy.
Only a few robots are using hearing capabilities: SAIL \cite{ZHANG}
uses one microphone to develop online audio-driven behaviors; ROBITA
\cite{matsusaka-tojo-kubota-furukawa-tamiya-hayata-nakano-kobayashi99}
uses two microphones to follow a conversation between two persons;
SIG \cite{nakadai-okuno-kitano2002,okuno-nakadai-kitano2002,nakadai-matsuura-okuno-kitano2003}
uses one pair of microphones to collect sound from the external world,
and another pair placed inside the head to collect internal sounds
(caused by motors) for noise cancellation; Sony SDR-4X has seven microphones;
a service robot uses eight microphones organised in a circular array
to do speech enhancement and recognition \cite{choi-kong-kim-bang2003}.
Even though robots are not limited to only two ears, they still have
not shown the capabilities of the human hearing sense.

We address the problem of isolating sound sources from the environment.
The human hearing sense is very good at focusing on a single source
of interest despite all kinds of interferences. We generally refer
to this ability as the \emph{cocktail party effect}, where a human
listener is able to follow a conversation even when several people
are speaking at the same time. For a mobile robot, it would mean being
able to separate all sound sources present in the environment at any
moment.

Working toward that goal, our interest in this paper is to describe
a two-step approach for performing sound source separation on a mobile
robot equipped with an array of eight low-cost microphones. The initial
step consists of a linear separation based on a simplified version
of the Geometric Source Separation approach proposed by Parra and
Alvino \cite{Parra2002Geometric} with a faster stochastic gradient
estimation and shorter time frames estimations. The second step is
a generalisation of beamformer post-filtering \cite{CohenArray2002,ValinICASSP2004}
for multiple sources and uses adaptive spectral estimation of background
noise and interfering sources to enhance the signal produced during
the initial separation. The novelty of this post-filter resides in
the fact that, for each source of interest, the noise estimate is
decomposed into stationary and  transient components assumed to be
due to leakage between the output channels of the initial separation
stage.

The paper is organised as follows. Section \ref{sec:System-overview}
gives an overview of the system. Section \ref{sec:LSS} presents the
linear separation algorithm and Section \ref{sec:Loudness-domain-spectral-attenuation}
describes the proposed post-filter. Results are presented in Section
\ref{sec:Results}, followed by the conclusion.

\section{System overview}

\label{sec:System-overview}

\begin{figure}
\includegraphics[width=1\columnwidth]{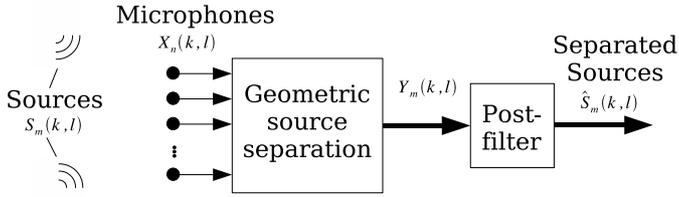}

\caption{Overview of the separation system\label{cap:Overview}}
\end{figure}

The proposed sound separation algorithm as shown in Figure \ref{cap:Overview}
is composed of three parts:
\begin{enumerate}
\item A microphone array;
\item A linear source separation algorithm (LSS) implemented as a variant
of the Geometric Source Separation (GSS) algorithm;
\item A multi-channel post-filter.
\end{enumerate}

The microphone array is composed of a number of omni-directional elements
mounted on the robot. The microphone signals are combined linearly
in a first-pass separation algorithm. The output of this initial separation
is then enhanced by a (non-linear) post-filter designed to optimally
attenuate the remaining noise and interference from other sources.

We assume that these sources are detected and localised by an algorithm
such as \cite{ValinICRA2004} (our approach is not specific to any
localisation algorithm). We also assume that sources may appear, disappear
or move at any time. It is thus necessary to maximise the adaptation
rate for both the LSS and the multi-channel post-filter. Mobile robotics
also imposes real-time constraints: the algorithmic delay must be
kept small and the complexity must be low enough for the data to be
processed in real-time on a conventional processor.

\begin{comment}
This fact imposes constraints on both the LSS and the post-filter. 

We are aware that the LSS algorithm is far from perfect (hence the
need for a post-filter) because of localisation accuracy, reverberation
and imperfect microphones (non-iden\-tical response). We design the
post-filter in such a way that any source separation algorithm (including
blind algorithms that do not require localisation of the sources)
can be used.
\end{comment}

\section{Linear Source Separation\label{sec:LSS}}

The LSS algorithm we propose in this section is based on the Geometric
Source Separation (GSS) approach proposed by Parra and Alvino \cite{Parra2002Geometric}.
Unlike the Linearly Constrained Minimum Variance (LCMV) beamformer
that minimises the output power subject to a distortionless constraint,
GSS explicitly minimises cross-talk, leading to faster adaptation.
The method is also interesting for use in the mobile robotics context
because it allows easy addition and removal of sources. Using some
approximations described in Subsection \ref{sub:Stochastic-Gradient-Adaptation},
it is also possible to implement separation with relatively low complexity
(i.e. complexity that grows linearly with the number of microphones).

\subsection{Geometric Source Separation\label{sub:Geometric-Source-Separation}}

The method operates in the frequency domain. Let $S_{m}(k,\ell)$
be the real (unknown) sound source $m$ at time frame $\ell$ and
for discrete frequency $k$. We denote as $\mathbf{s}(k,\ell)$ the
vector corresponding to the sources $S_{m}(k,\ell)$ and matrix $\mathbf{A}(k)$
is the transfer function leading from the sources to the microphones.
The signal received at the microphones is thus given by:
\begin{equation}
\mathbf{x}(k,\ell)=\mathbf{A}(k)\mathbf{s}(k,\ell)+\mathbf{n}(k,\ell)\label{eq:GSS_mixing}
\end{equation}
where $\mathbf{n}(k,\ell)$ is the non-coherent background noise received
at the microphones. The matrix $\mathbf{A}(k)$ can be estimated using
the result of a sound localisation algorithm. Assuming that all transfer
functions have unity gain, the elements of $\mathbf{A}(k)$ can be
expressed as:
\begin{equation}
a_{ij}(k)=e^{-\jmath2\pi k\delta_{ij}}\label{eq:GSS_a_ij}
\end{equation}
where $\delta_{ij}$ is the time delay (in samples) to reach microphone
$i$ from source $j$. 

The separation result is then defined as $\mathbf{y}(k,\ell)=\mathbf{W}(k,\ell)\mathbf{x}(k,\ell)$,
where $\mathbf{W}(k,\ell)$ is the separation matrix that must be
estimated. This is done by providing two constraints (the index $\ell$
is omitted for the sake of clarity):
\begin{enumerate}
\item Decorrelation of the separation algorithm outputs, expressed as $\mathbf{R}_{\mathbf{yy}}(k)-\mathrm{diag}\left[\mathbf{R}_{\mathbf{yy}}(k)\right]=\mathbf{0}$\footnote{Assuming non-stationary sources, second order statistics are sufficient
for ensuring independence of the separated sources.}.
\item The geometric constraint $\mathbf{W}(k)\mathbf{A}(k)=\mathbf{I}$,
which ensures unity gain in the direction of the source of interest
and places zeros in the direction of interferences.
\end{enumerate}
In theory, constraint 2) could be used alone for separation (the method
is referred to as LS-C2 in \cite{Parra2002Geometric}), but in practice,
the method does not take into account reverberation or errors in localisation.
It is also subject to instability if $\mathbf{A}(k)$ is not invertible
at a specific frequency. When used together, constraints 1) and 2)
are too strong. For this reason, we propose ``soft'' constraints
that are a combination of 1) and 2) in the context of a gradient descent
algorithm.

Two cost functions are created by computing the square of the error
associated with constraints 1) and 2). These cost functions are respectively
defined as:
\begin{eqnarray}
J_{1}(\mathbf{W}(k)) & = & \left\Vert \mathbf{R}_{\mathbf{yy}}(k)-\mathrm{diag}\left[\mathbf{R}_{\mathbf{yy}}(k)\right]\right\Vert ^{2}\label{eq:Cost_J1}\\
J_{2}(\mathbf{W}(k)) & = & \left\Vert \mathbf{W}(k)\mathbf{A}(k)-\mathbf{I}\right\Vert ^{2}\label{eq:Cost_J2}
\end{eqnarray}
where the matrix norm is defined as $\left\Vert \mathbf{M}\right\Vert ^{2}=\mathrm{trace}\left[\mathbf{M}\mathbf{M}^{H}\right]$
and is equal to the sum of the square of all elements in the matrix.
The gradient of the cost functions with respect to $\mathbf{W}(k)$
is equal to \cite{Parra2002Geometric}:
\begin{eqnarray}
\frac{\partial J_{1}(\mathbf{W}(k))}{\partial\mathbf{W}^{*}(k)} & = & 4\mathbf{E}(k)\mathbf{W}(k)\mathbf{R}_{\mathbf{xx}}(k)\label{eq:Grad_J1}\\
\frac{\partial J_{2}(\mathbf{W}(k))}{\partial\mathbf{W}^{*}(k)} & = & 2\left[\mathbf{W}(k)\mathbf{A}(k)-\mathbf{I}\right]\mathbf{A}(k)\label{eq:Grad_J2}
\end{eqnarray}
where $\mathbf{E}(k)=\mathbf{R}_{\mathbf{yy}}(k)-\mathrm{diag}\left[\mathbf{R}_{\mathbf{yy}}(k)\right]$.

The separation matrix $\mathbf{W}(k)$ is then updated as follows:
\begin{equation}
\mathbf{W}^{n+1}(k)=\mathbf{W}^{n}(k)-\mu\!\left[\alpha(k)\frac{\partial J_{1}(\mathbf{W}(k))}{\partial\mathbf{W}^{*}(k)}\!+\!\frac{\partial J_{2}(\mathbf{W}(k))}{\partial\mathbf{W}^{*}(k)}\right]\label{eq:W_update}
\end{equation}
where $\alpha(f)$ is an energy normalisation factor equal to $\left\Vert \mathbf{R}_{\mathbf{xx}}(k)\right\Vert ^{-2}$
and $\mu$ is the adaptation rate.

\subsection{Stochastic Gradient Adaptation\label{sub:Stochastic-Gradient-Adaptation}}

The difference between our algorithm and the original GSS algorithm
described in \cite{Parra2002Geometric} is that instead of estimating
the correlation matrices $\mathbf{R}_{\mathbf{xx}}(k)$ and $\mathbf{R}_{\mathbf{yy}}(k)$
on several seconds of data, our approach uses instantaneous estimations.
This is analogous to the approximation made in the Least Mean Square
(LMS) adaptive filter \cite{HaykinAFT}. We thus assume that:
\begin{eqnarray}
\mathbf{R}_{\mathbf{xx}}(k) & = & \mathbf{x}(k)\mathbf{x}(k)^{H}\label{eq:Rxx_instant}\\
\mathbf{R}_{\mathbf{yy}}(k) & = & \mathbf{y}(k)\mathbf{y}(k)^{H}\label{eq:Ryy_instant}
\end{eqnarray}

It is then possible to rewrite the gradient $\frac{\partial J_{1}(\mathbf{W}(k))}{\partial\mathbf{W}^{*}(k)}$
as:
\begin{equation}
\frac{\partial J_{1}(\mathbf{W}(k))}{\partial\mathbf{W}^{*}(k)}=4\left[\mathbf{E}(k)\mathbf{W}(k)\mathbf{x}(k)\right]\mathbf{x}(k)^{H}\label{eq:Grad_J1_instant}
\end{equation}
which only requires matrix-by-vector products, greatly reducing the
complexity of the algorithm. The normalisation factor $\alpha(k)$
can also be simplified as $\left[\left\Vert \mathbf{x}(k)\right\Vert ^{2}\right]^{-2}$.
From this work, the instantaneous estimation of the correlation has
not shown any reduction in accuracy and furthermore eases real-time
integration.

\subsection{Initialisation}

The fact that sources can appear or disappear at any time imposes
constraints on the initialisation of the separation matrix $\mathbf{W}(k)$.
The initialisation must provide the following:
\begin{itemize}
\item The initial weights for a new source;
\item Acceptable separation (before adaptation).
\end{itemize}
Furthermore, when a source appears or disappears, other sources must
be unaffected.

One easy way to satisfy both constraints is to initialise the column
of $\mathbf{W}(k)$ corresponding to the new source $m$ as:
\begin{equation}
w_{m,i}(k)=\frac{a_{i,m}(k)}{N}\label{eq:GSS_I1}
\end{equation}

This initialisation is equivalent to a delay-and-sum beamformer, and
is referred to as the I1 initialisation method in \cite{Parra2002Geometric}.

\section{Multi-channel post-filter}

\label{sec:Loudness-domain-spectral-attenuation}

\begin{figure}
\includegraphics[width=1\columnwidth,keepaspectratio]{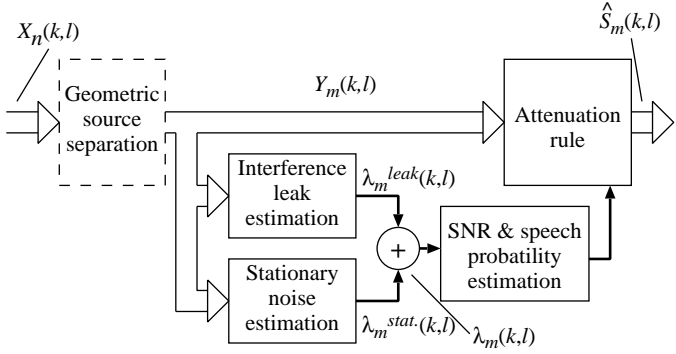}

\caption{Overview of the post-filter.\protect \\
$X_{n}(k,\ell),n=0\ldots N-1$: Microphone inputs, $Y_{m}(k,\ell),\:m=0\ldots M-1$:
Inputs to the post-filter, $\hat{S}_{m}(k,\ell)=G_{m}(k,\ell)Y_{m}(k,\ell),\:m=0\ldots M-1$:
Post-filter outputs.\label{cap:Overview-Post-filter}}
\end{figure}

In order to enhance the output of the GSS algorithm presented in Section
\ref{sec:LSS}, we derive a frequency-domain post-filter that is based
on the optimal estimator originally proposed by Ephraim and Malah
\cite{EphraimMalah1984,EphraimMalah1985}. Several approaches to microphone
array post-filtering have been proposed in the past. Most of these
post-filters address reduction of stationary background noise \cite{zelinski-1988,mccowan-icassp2002}.
Recently, a multi-channel post-filter taking into account non-stationary
interferences was proposed by Cohen \cite{CohenArray2002}. The novelty
of our approach resides in the fact that, for a given channel output
of the GSS, the transient components of the corrupting sources is
assumed to be due to leakage from the other channels during the GSS
process. Furthermore, for a given channel, the stationary and the
transient components are combined into a single noise estimator used
for noise suppression, as shown in Figure \ref{cap:Overview-Post-filter}. 

For this post-filter, we consider that all interferences (except the
background noise) are localised (detected by the localisation algorithm)
sources and we assume that the leakage between channels is constant.
This leakage is due to reverberation, localisation error, differences
in microphone frequency responses, near-field effects, etc.

Section \ref{sub:Noise-estimation} describes the estimation of noise
variances that are used to compute the weighting function $G_{m}$
by which the outputs $Y_{m}$ of the LSS is multiplied to generate
a cleaned signal whose spectrum is denoted $\hat{S}_{m}$.

\subsection{Noise estimation\label{sub:Noise-estimation}}

The noise variance estimation $\lambda_{m}(k,\ell)$ is expressed
as:
\begin{equation}
\lambda_{m}(k,\ell)=\lambda_{m}^{stat.}(k,\ell)+\lambda_{m}^{leak}(k,\ell)\label{eq:noise_estim}
\end{equation}
where $\lambda_{m}^{stat.}(k,\ell)$ is the estimate of the stationary
component of the noise for source $m$ at frame $\ell$ for frequency
$k$, and $\lambda_{m}^{leak}(k,\ell)$ is the estimate of source
leakage.

We compute the stationary noise estimate $\lambda_{m}^{stat.}(k,\ell)$
using the Minima Controlled Recursive Average (MCRA) technique proposed
by Cohen \cite{CohenNonStat2001}. 

To estimate $\lambda_{m}^{leak}$ we assume that the interference
from other sources is reduced by a factor $\eta$ (typically $-10\:\mathrm{dB}\leq\eta\leq-5\:\mathrm{dB}$)
by the separation algorithm (LSS). The leakage  estimate is thus expressed
as:
\begin{equation}
\lambda_{m}^{leak}(k,\ell)=\eta\sum_{i=0,i\neq m}^{M-1}Z_{i}(k,\ell)\label{eq:leak_estim}
\end{equation}
where $Z_{m}(k,\ell)$ is the smoothed spectrum of the $m^{th}$ source,
$Y_{m}(k,\ell)$, and is recursively defined (with $\alpha_{s}=0.7$)
as:
\begin{equation}
Z_{m}(k,\ell)=\alpha_{s}Z_{m}(k,\ell-1)+(1-\alpha_{s})Y_{m}(k,\ell)\label{eq:smoothed_spectrum}
\end{equation}

\subsection{Suppression rule in the presence of speech}

We now derive the suppression rule under $H_{1}$, the hypothesis
that speech is present. From here on, unless otherwise stated, the
$m$ index and the $\ell$ arguments are omitted for clarity and the
equations are given for each $m$ and for each $\ell$.

The proposed noise suppression rule is based on minimum mean-square
error (MMSE) estimation of the spectral amplitude in the loudness
domain, $\left|X(k)\right|^{1/2}$. The choice of the loudness domain
over the spectral amplitude \cite{EphraimMalah1984} or log-spectral
amplitude \cite{EphraimMalah1985} is motivated by better results
obtained using this technique, mostly when dealing with speech presence
uncertainty (Section \ref{sec:Optimal-gain-modification}).

The loudness-domain amplitude estimator is defined by:
\begin{equation}
\hat{A}(k)=\left(E\left[\left|S(k)\right|^{\alpha}\left|Y(k)\right.\right]\right)^{\frac{1}{\alpha}}=G_{H_{1}}(k)\left|Y(k)\right|\label{eq:amplitude_estim}
\end{equation}
where $\alpha=1/2$ for the loudness domain and $G_{H_{1}}(k)$ is
the spectral gain assuming that speech is present. 

The spectral gain for arbitrary $\alpha$ is derived from Equation
13 in \cite{EphraimMalah1985}:
\begin{equation}
G_{H_{1}}(k)=\frac{\sqrt{\upsilon(k)}}{\gamma(k)}\left[\Gamma\left(1+\frac{\alpha}{2}\right)M\left(-\frac{\alpha}{2};1;-\upsilon(k)\right)\right]^{\frac{1}{\alpha}}\label{eq:Gain_H1}
\end{equation}
where $M(a;c;x)$ is the confluent hypergeometric function, $\gamma(k)\triangleq\left|Y(k)\right|^{2}/\lambda(k)$
and $\xi(k)\triangleq E\left[\left|S(k)\right|^{2}\right]/\lambda(k)$
are respectively the \emph{a posteriori} SNR and the \emph{a priori}
SNR. We also have $\upsilon(k)\triangleq\gamma(k)\xi(k)/\left(\xi(k)+1\right)$
\cite{EphraimMalah1984}.

The \emph{a priori} SNR $\xi(k)$ is estimated recursively as:
\begin{eqnarray}
\hat{\xi}(k,l) & = & \alpha_{p}G_{H_{1}}^{2}(k,\ell-1)\gamma(k,\ell-1)\nonumber \\
 & + & (1-\alpha_{p})\max\left\{ \gamma(k,\ell)-1,0\right\} \label{eq:xi_decision_directed}
\end{eqnarray}
using the modifications proposed in \cite{CohenNonStat2001} to take
into account speech presence uncertainty.

\subsection{Optimal gain modification under speech presence uncertainty}

\label{sec:Optimal-gain-modification}In order to take into account
the probability of speech presence, we derive the estimator for the
loudness domain:
\begin{equation}
\hat{A}(k)=\left(E\left[\left.A^{\alpha}(k)\right|Y(k)\right]\right)^{\frac{1}{\alpha}}\label{eq:optimal-gain}
\end{equation}

Considering $H_{1}$, the hypothesis of speech presence for source
$m$, and $H_{0}$, the hypothesis of speech absence, we obtain:
\begin{eqnarray}
E\!\left[\left.\!A^{\alpha}(k)\right|\!Y\!(k)\right] & = & p(k)E\left[\left.A^{\alpha}(k)\right|H_{1},Y(k)\right]\nonumber \\
 & + & \left[1\!-\!p(k)\right]\!E\!\left[\left.\!A^{\alpha}(k)\right|\!H_{0},\!Y\!(k)\right]\label{eq:optimal-gain2}
\end{eqnarray}
where $p(k)$ is the probability of speech at frequency $k$.

The optimally modified gain is thus given by:
\begin{equation}
G(k)=\left[p(k)G_{H_{1}}^{\alpha}(k)+(1-p(k))G_{min}^{\alpha}\right]^{\frac{1}{\alpha}}\label{eq:optimal_gain3}
\end{equation}
where $G_{H_{1}}(k)$ is defined in (\ref{eq:Gain_H1}), and $G_{min}$
is the minimum gain allowed when speech is absent. Unlike the log-amplitude
case, it is possible to set $G_{min}=0$ without running into problems.
For $\alpha=1/2$, this leads to:
\begin{equation}
G(k)=p^{2}(k)G_{H_{1}}(k)\label{eq:optimal_gain_final}
\end{equation}

Setting $G_{min}=0$ means that there is no arbitrary limit on attenuation.
Therefore, when the signal is certain to be non-speech, the gain can
tend toward zero. This is especially important when the interference
is also speech since, unlike stationary noise, residual babble noise
always results in musical noise.

The probability of speech presence is computed as:
\begin{equation}
p(k)=\left\{ 1+\frac{\hat{q}(k)}{1-\hat{q}(k)}\left(1+\xi(k)\right)\exp\left(-\upsilon(k)\right)\right\} ^{-1}\label{eq:def_speech_prob}
\end{equation}
where $\hat{q}(k)$ is the \emph{a priori} probability of speech presence
for frequency $k$ and is defined as:
\begin{equation}
\hat{q}(k)=1-P_{local}(k)P_{global}(k)P_{frame}\label{eq:def_q_a_priori}
\end{equation}
where $P_{local}(k)$, $P_{global}(k)$ and $P_{frame}$ are defined
in \cite{CohenNonStat2001} and correspond respectively to a speech
measurement on the current frame for a local frequency window, a larger
frequency and for the whole frame.

\subsection{Initialisation}

When a new source appears, post-filter state variables need to be
initialised. Most of these variables may safely be set to zero. The
exception is $\lambda_{m}^{stat.}\left(k,\ell_{0}\right)$, the initial
stationary noise estimation for source $m$. The MCRA algorithm requires
several seconds to produce its first estimate for source $m$, so
it is necessary to find another way to estimate the background noise
until a better estimate is available. This initial estimate is thus
computed using noise estimations at the microphones. Assuming the
delay-and-sum initialisation of the weights from Equation \ref{eq:GSS_I1},
the initial background noise estimate is thus:
\begin{equation}
\lambda_{m}^{stat.}\left(k,\ell_{0}\right)=\frac{1}{N^{2}}\sum_{n=0}^{N-1}\sigma_{x_{n}}^{2}\left(k\right)\label{eq:noise_init}
\end{equation}
where $\sigma_{x_{n}}^{2}\left(k\right)$ is the noise estimation
for microphone $n$.

\section{Results}

\begin{figure}
\begin{center}\includegraphics[width=0.9\columnwidth]{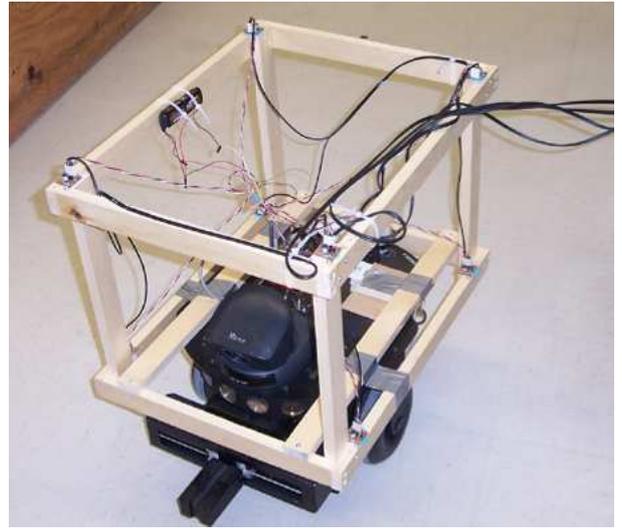}\end{center}

\caption{Pioneer 2 robot with an array of eight microphones\label{cap:Pioneer2}}
\end{figure}

\label{sec:Results}Our system is evaluated on a Pioneer 2 robot,
on which an array of eight microphones is installed. In order to test
the system, three voices (two female, one male) were recorded separately,
in a quiet environment. The background noise was recorded on the robot
and includes the room ventilation and the internal robot fans. All
four signals were recorded using the same microphone array and subsequently
mixed together. This procedure was required in order to compute the
distance measures (such as SNR) presented in this section. It is worth
noting that although the signals were mixed artificially, the result
still represents real conditions with background noise, interfering
sources, and reverberation.

In evaluating our source separation system, we use the conventional
signal-to-noise ratio (SNR) and the log spectral distortion (LSD),
that is defined as:
\begin{equation}
LSD=\frac{1}{L}\sum_{\ell=0}^{L-1}\left[\frac{1}{K}\sum_{k=0}^{K-1}\left(10\log_{10}\frac{\left|S(k,\ell)\right|^{2}+\epsilon}{\left|\hat{S}(k,\ell)\right|^{2}+\epsilon}\right)^{2}\right]^{\frac{1}{2}}\label{eq:def_LSD}
\end{equation}
where $L$ is the number of frames, $K$ is the number of frequency
bins and $\epsilon$ is meant to prevent extreme values for spectral
regions of low energy.

\begin{table}
\caption{Signal-to-noise ratio (SNR) for each of the three separated sources.\label{cap:Signal-to-noise-ratio-(SNR)}}

\begin{center}%
\begin{tabular}{|c|c|c|c|}
\hline 
SNR (dB) & female 1 & female 2 & male 1\tabularnewline
\hline 
\hline 
Microphone inputs & -1.8 & -3.7 & -5.2\tabularnewline
\hline 
Delay-and-sum & 7.3 & 4.4 & -1.2\tabularnewline
\hline 
GSS & 9.0 & 6.0 & 3.7\tabularnewline
\hline 
GSS+single channel & 9.9 & 6.9 & 4.5\tabularnewline
\hline 
GSS+multi-channel & 12.1 & 9.5 & 9.4\tabularnewline
\hline 
\end{tabular}\end{center}

\caption{Log-spectral distortion (LSD) for each of the three separated sources.\label{cap:Log-spectral-distortion-(LSD)}}

\begin{center}%
\begin{tabular}{|c|c|c|c|}
\hline 
LSD (dB) & female 1 & female 2 & male 1\tabularnewline
\hline 
\hline 
Microphone inputs & 17.5 & 15.9 & 14.8\tabularnewline
\hline 
Delay-and-sum & 15.8 & 15.0 & 15.1\tabularnewline
\hline 
GSS & 15.0 & 14.2 & 14.2\tabularnewline
\hline 
GSS+single channel & 9.7 & 9.5 & 10.4\tabularnewline
\hline 
GSS+multi-channel & 6.5 & 6.8 & 7.4\tabularnewline
\hline 
\end{tabular}\end{center}
\end{table}

Tables \ref{cap:Signal-to-noise-ratio-(SNR)} and \ref{cap:Log-spectral-distortion-(LSD)}
compare the results obtained for different configurations: unprocessed
microphone inputs, delay-and-sum algorithm, GSS algorithm, GSS algorithm
with single-channel post-filter, and GSS algorithm with multi-channel
post-filter (proposed). It is worth noting that the delay-and-sum
algorithm corresponds to the initial value of the separation matrix
provided to our algorithm. While it is clear that GSS performs better
than delay-and-sum, the latter still provides acceptable separation
capabilities. These results also show that our multi-channel post-filter
provides a significant improvement over both the single-channel post-filter
and plain GSS.

\begin{figure*}
a)\includegraphics[width=0.4\paperwidth,height=0.2\paperwidth]{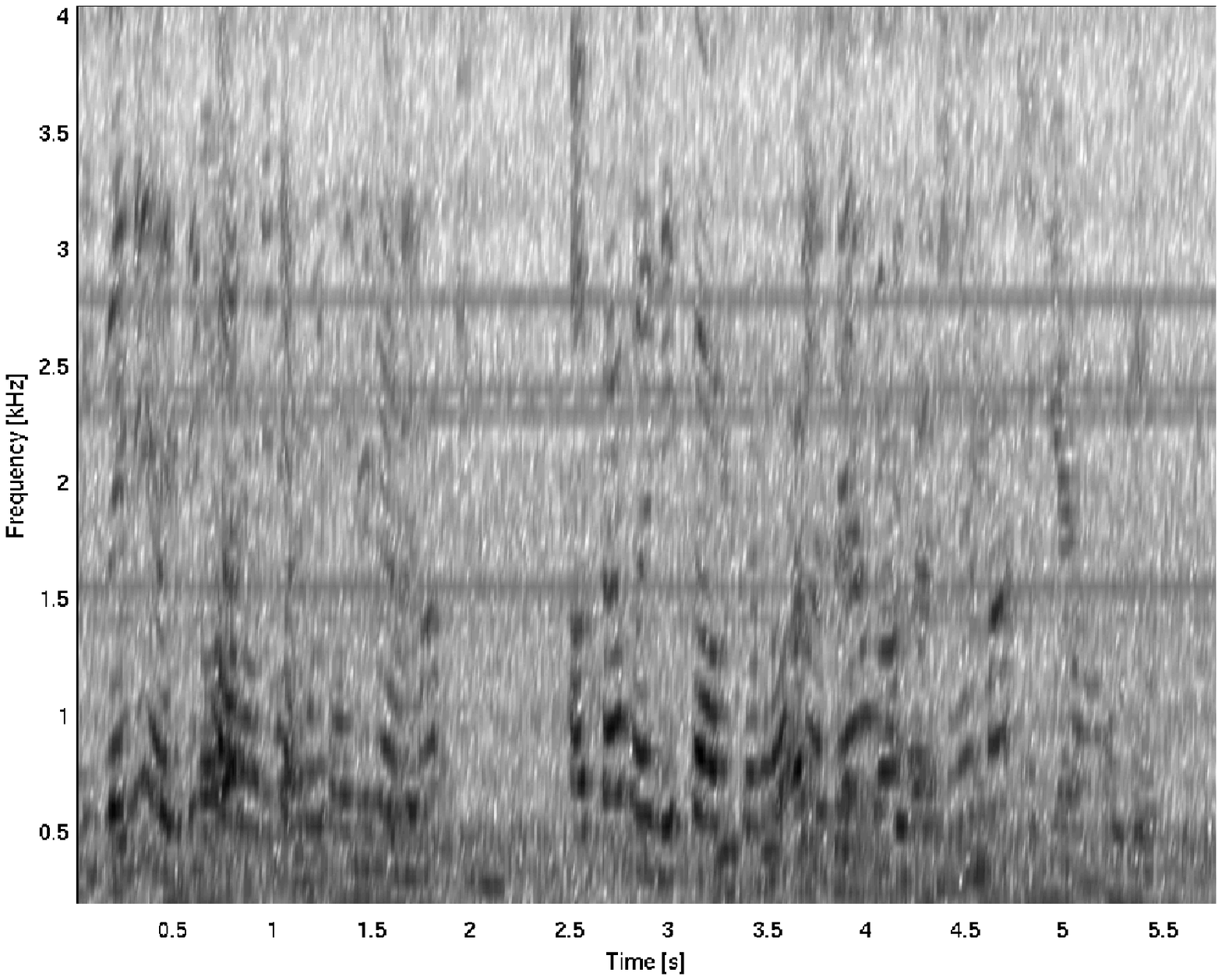}b)\includegraphics[width=0.4\paperwidth,height=0.2\paperwidth]{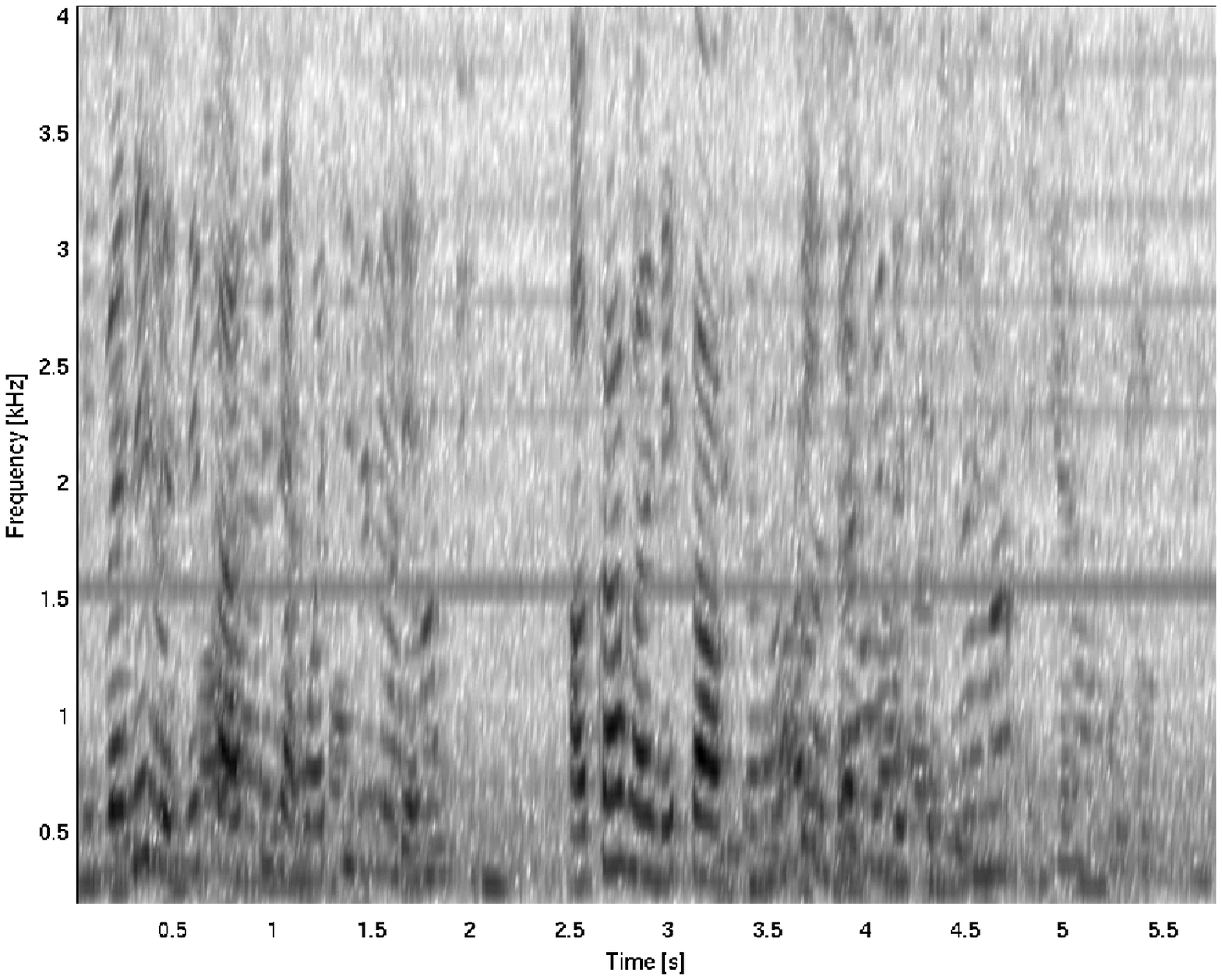}

c)\includegraphics[width=0.4\paperwidth,height=0.2\paperwidth]{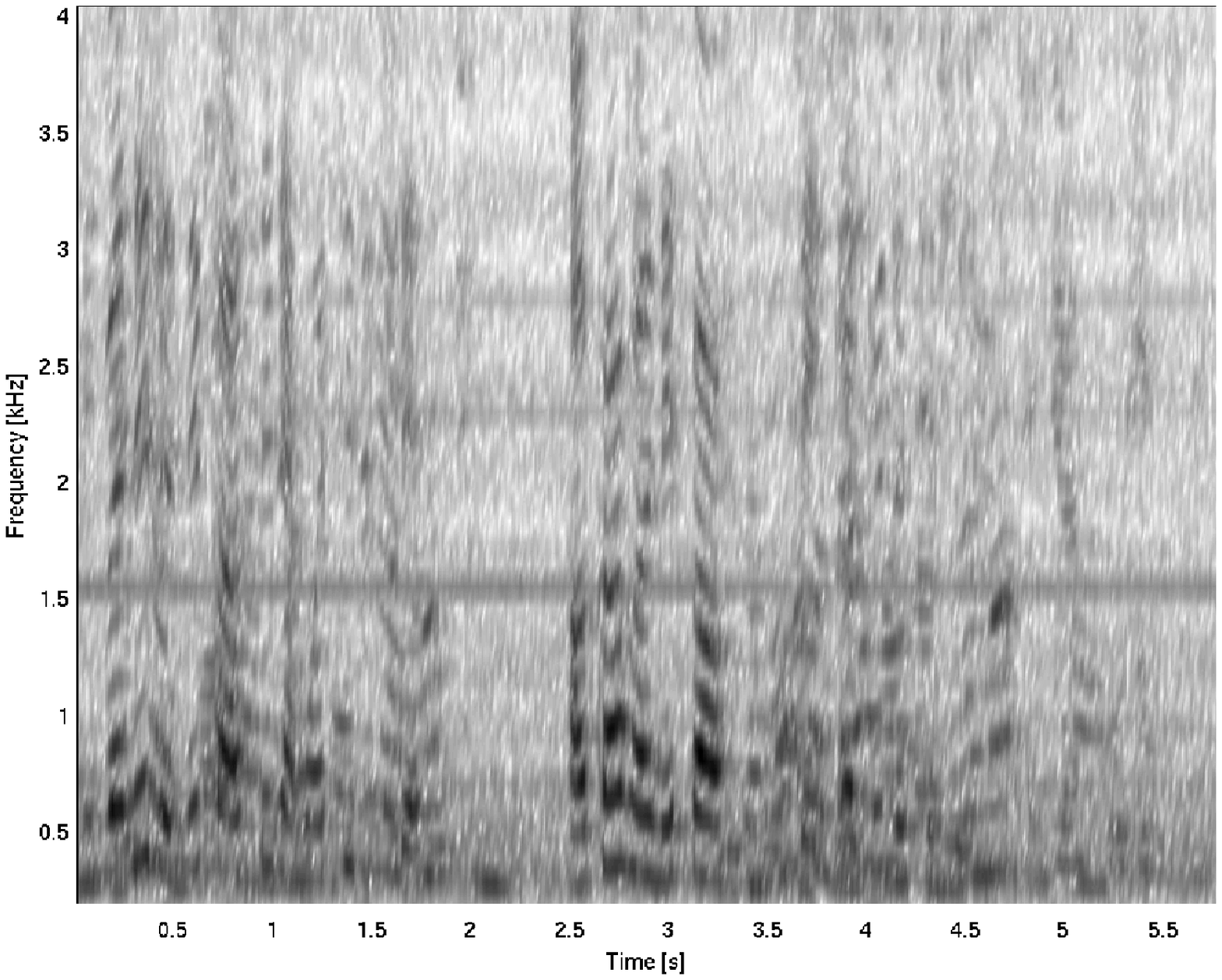}d)\includegraphics[width=0.4\paperwidth,height=0.2\paperwidth]{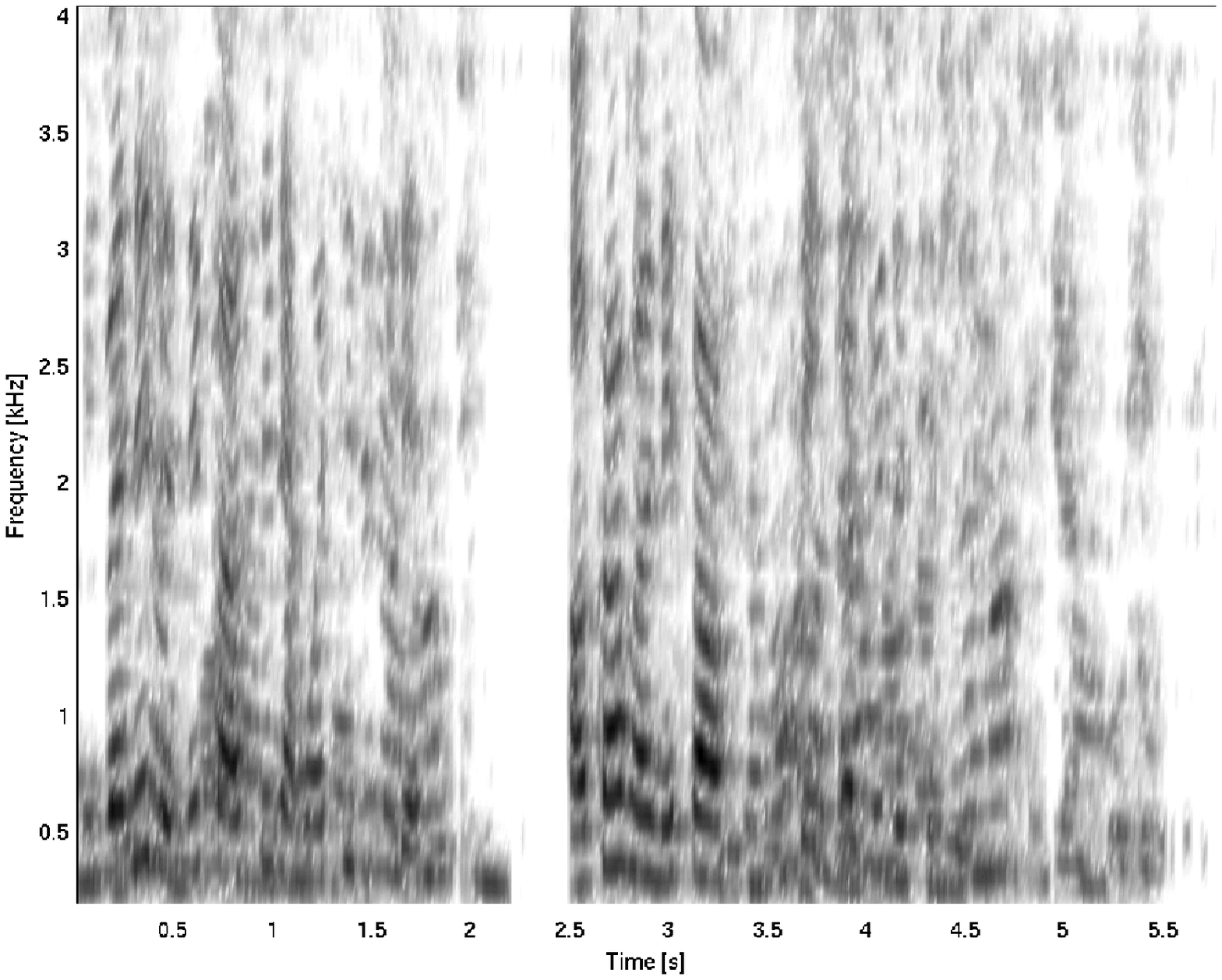}

e)\includegraphics[width=0.4\paperwidth,height=0.2\paperwidth]{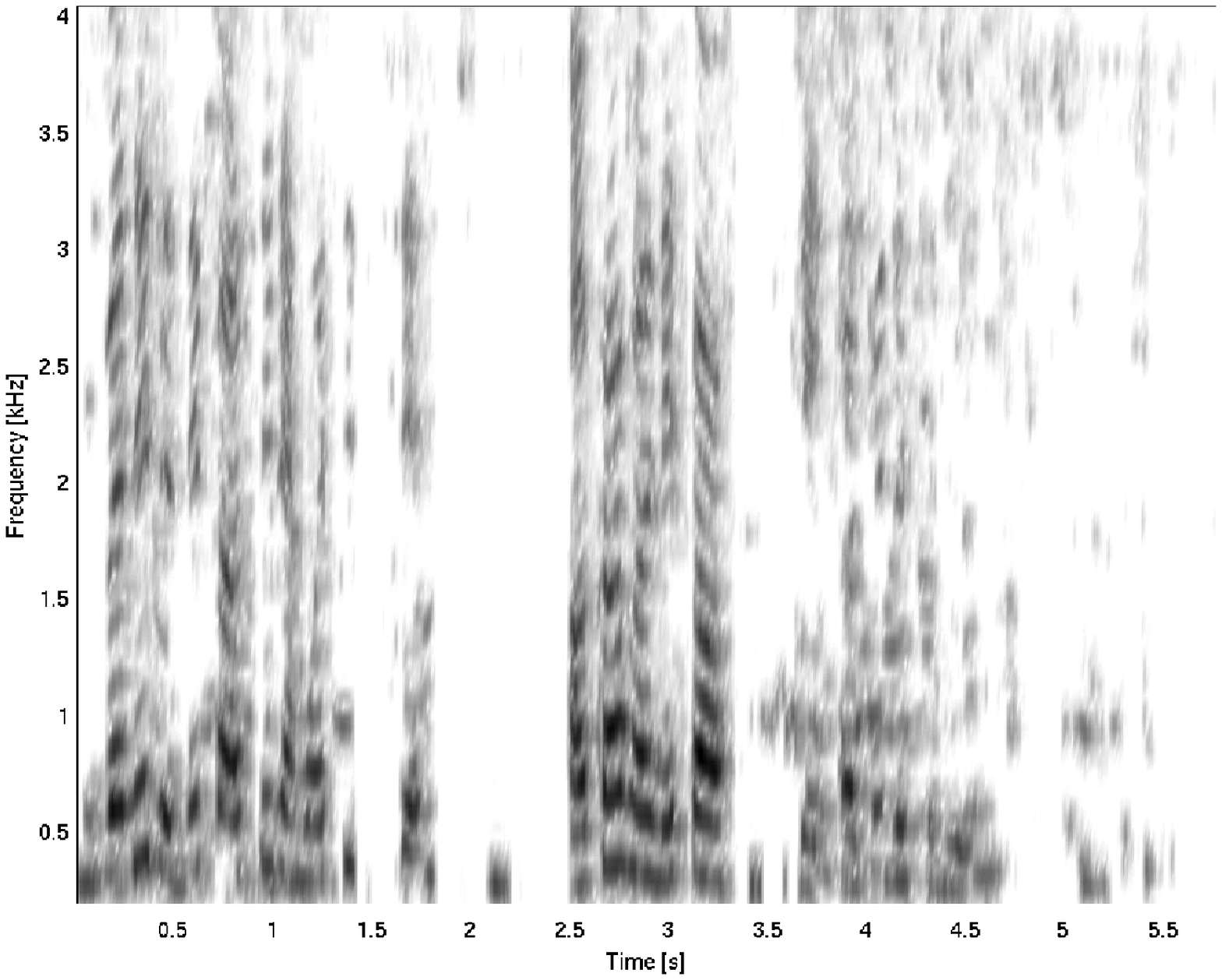}f)\includegraphics[width=0.4\paperwidth,height=0.2\paperwidth]{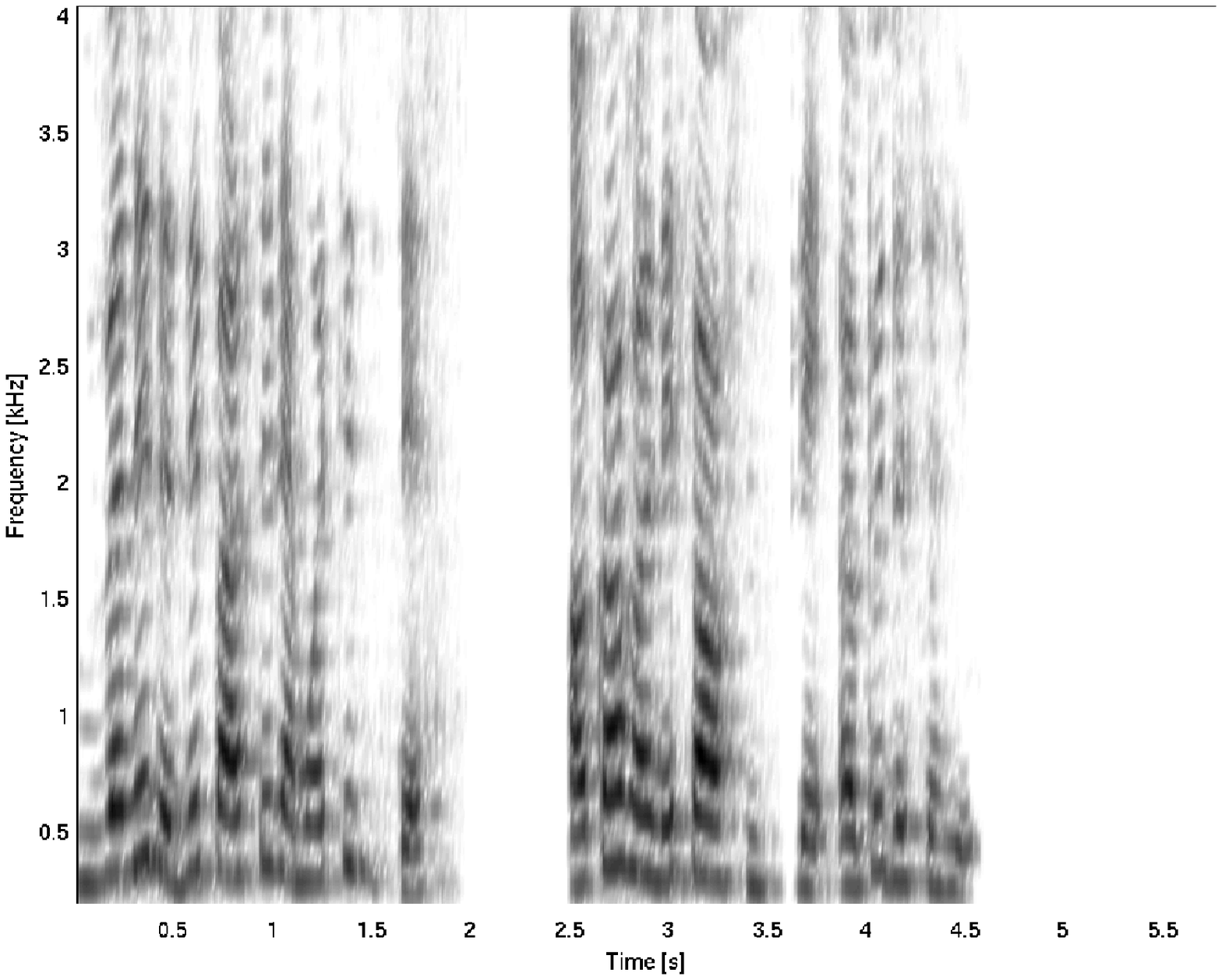}

\caption{Spectrograms for separation of first source (female voice): a) signal
at one microphone, b) delay-and-sum beamformer, c) GSS output, d)
GSS with single-channel post-filter, e) GSS with multi-channel post-filter,
f) reference (clean) signal. \label{cap:Spectrogram-for-separation}}
\end{figure*}

\begin{figure}
\includegraphics[width=1\columnwidth]{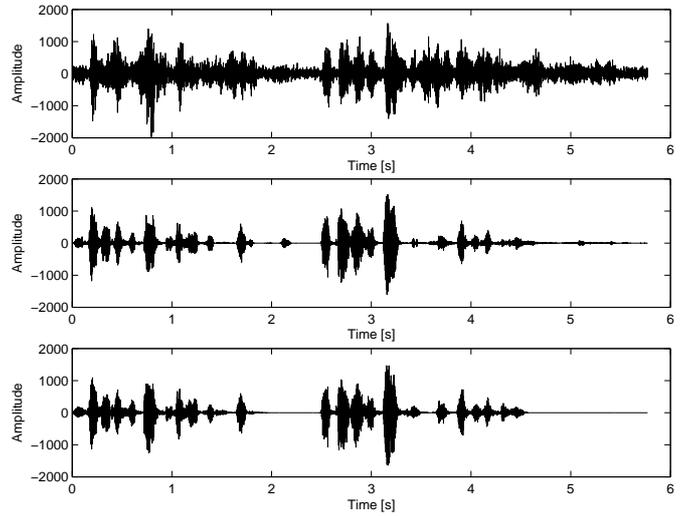}

\caption{Signal amplitude for separation of first source (female voice). top:
signal at one microphone. middle: system output. bottom: reference
(clean) signal.\label{cap:Signal-amplitude}}
\end{figure}

The signals amplitude for the first source (female) are shown in Figure
\ref{cap:Signal-amplitude} and the spectrograms are shown in Figure
\ref{cap:Spectrogram-for-separation}. Even though the task involves
non-stationary interference with the same frequency content as the
signal of interest, we observe that our post-filter (unlike the single-channel
post-filter) is able to remove most of the interference, while not
causing excessive distortion to the signal of interest. Informal subjective
evaluation has confirmed that the post-filter has a positive impact
on both quality and intelligibility of the speech\footnote{Audio signals and spectrograms for all three sources are available
at: \texttt{http://www.speex.org/\textasciitilde{}jm/phd/separation/}}.

\section{Conclusion}

\label{sec:Discussion}In this paper we describe a microphone array
linear source separator and a post-filter in the context of multiple
and simultaneous sound sources. The linear source separator is based
on a simplification of the geometric source separation algorithm that
performs instantaneous estimation of the correlation matrix $\mathbf{R}_{\mathbf{xx}}(k)$.
The post-filter is based on a loudness-domain MMSE estimator in the
frequency domain with a noise estimate that is computed as the sum
of a stationary noise estimate and an estimation of leakage from the
geometric source separation algorithm. The proposed post-filter is
also sufficiently general to be used in addition to most linear source
separation algorithms. 

Experimental results show a reduction in log spectral distortion of
up to $11\:\mathrm{dB}$ and an increase of the signal-to-noise ratio
of $14\:\mathrm{dB}$ compared to the noisy signal inputs. Preliminary
perceptive test and visual inspection of spectrograms show us that
the distortions introduced by the system are acceptable to most listeners.

A possible next step for this work would consist of directly optimizing
the separation results for speech recognition accuracy. Also, a possible
improvement to the algorithm would be to derive a method that automatically
adapts the leakage coefficient $\eta$ to track the leakage of the
GSS algorithm.

\begin{comment}
Jean-Marc: idees d'améliorations possibles du système?

A possible improvement to the algorithm would be to derive a method
that automatically adapts the leakage factor $\eta$ to track the
leakage of an adaptive LSS algorithm.
\end{comment}

\section*{Acknowledgment}

Fran\c{c}ois Michaud holds the Canada Research Chair (CRC) in Mobile
Robotics and Autonomous Intelligent Systems. This research is supported
financially by the CRC Program, the Natural Sciences and Engineering
Research Council of Canada (NSERC) and the Canadian Foundation for
Innovation (CFI). Special thanks to Dominic L\'{e}tourneau and Serge
Caron for their help in this work.

\bibliographystyle{IEEEtran}
\bibliography{iros,BiblioAudible}

\end{document}